\documentclass[10pt,twocolumn,letterpaper]{article}

\usepackage{iccv}              

\usepackage{amsmath} 
\usepackage{amssymb}

\usepackage{times} 
\usepackage{bbm} 
\usepackage{bm}

\usepackage{graphicx} 

\usepackage{xcolor}
\usepackage{color}
\definecolor{mygray}{gray}{.9}
\usepackage{colortbl} 
\usepackage{multirow}
\makeatletter
\newcommand{\thickhline}{%
    \noalign {\ifnum 0=`}\fi \hrule height 1pt
    \futurelet \reserved@a \@xhline
}
\makeatother

\usepackage{caption} 

\usepackage[export]{adjustbox} 

\usepackage{algorithm}
\usepackage{listings} 
\usepackage{array}
\usepackage{makecell}

\definecolor{iccvblue}{rgb}{0.21,0.49,0.74}

\usepackage[pagebackref=true,breaklinks=true,colorlinks=true,bookmarks=false]{hyperref}

\def\paperID{2840} 
\def\confName{ICCV}
\def\confYear{2025}

\begin{document}

\title{3D Gaussian Map with Open-Set Semantic Grouping \\ for Vision-Language Navigation}
\author{{Jianzhe Gao
\quad Rui Liu
\quad Wenguan Wang\thanks{Corresponding author: Wenguan Wang.}}\\
 \small{The State Key Lab of Brain-Machine Intelligence, Zhejiang University}
\\
\small \url{https://github.com/Gaozzzz/3D-Gaussian-Map-VLN}}
\maketitle


\begin{abstract}
Vision-language navigation (VLN) requires an agent to traverse complex 3D environments based on natural language instructions, necessitating a thorough scene understanding. While existing works equip agents with various scene representations to enhance spatial awareness, they often neglect the complex 3D geometry and rich semantics in VLN scenarios, limiting the ability to generalize across diverse and unseen environments. To address these challenges, this work proposes a \textbf{3D Gaussian Map} that represents the environment as a set of differentiable 3D Gaussians and accordingly develops a navigation strategy for VLN. Specifically, \textbf{Egocentric Scene Map} is constructed online by initializing 3D Gaussians from sparse pseudo-lidar point clouds, providing informative geometric priors for scene understanding. Each Gaussian primitive is further enriched through \textbf{Open-Set Semantic Grouping} operation, which groups 3D Gaussians based on their membership in object instances or stuff categories within the open world, resulting in a unified 3D Gaussian Map. Building on this map, \textbf{Multi-Level Action Prediction} strategy, which combines spatial-semantic cues at multiple granularities, is designed to assist agents in decision-making. Extensive experiments conducted on three public benchmarks (\textit{i.e.,} R2R, R4R, and REVERIE) validate the effectiveness of our method.

\end{abstract}
\section{Introduction}
\label{sec:intro}
\begin{figure}[t]
	\begin{center}
		\includegraphics[width=0.99\linewidth]{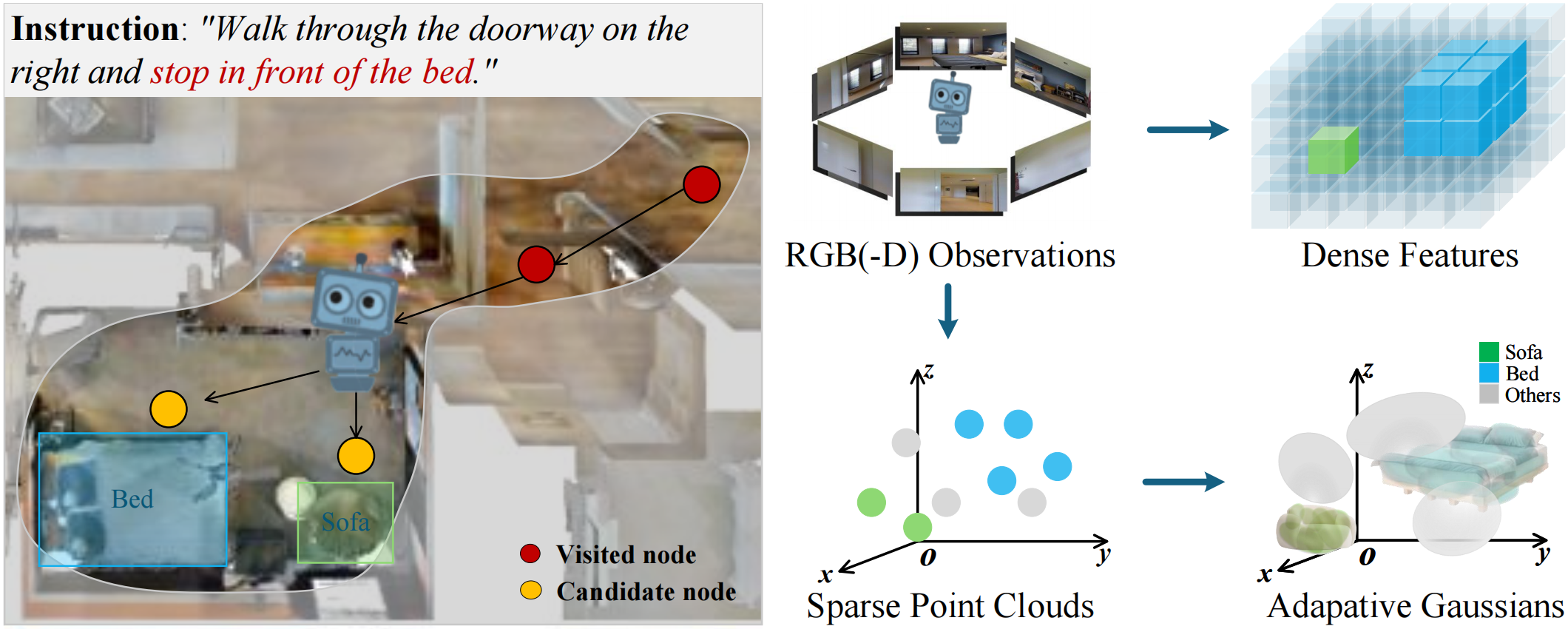}
	\end{center}
	\vspace{-15pt}
	\captionsetup{font=small}
	\caption{\small{
\textbf{Dense Features \textit{vs} 3D Gaussians.} Recent VLN methods~\cite{liu2023bird,an2023bevbert,liu2024volumetric,wang2024lookahead} rely on dense sampling to construct scene maps, which often leads to redundant representations and high computational costs. In contrast, our method introduces a set of sparse and adaptive 3D Gaussians to model the 3D scene, efficiently capturing spatial structures and integrating open-set semantics.}}
	\label{fig_introduce}
	\vspace{-6pt}
\end{figure}

Vision-and-Language Navigation (VLN) is a fundamental task in embodied AI, requiring an agent to interpret natural language instructions for navigating through diverse 3D environments~\cite{liu2024vision}. A core aspect of this task lies in improving the agent's perception and understanding of its environment, enabling it to reason about spatial structures, adapt to varying situations, and make informed decisions~\cite{zhong2024unrealzoo,driess2022reinforcement}.

Early VLN approaches~\cite{fried2018speaker,AndersonWTB0S0G18,tan2019learning,wang2022towards,gao2023room} primarily rely on sequence-to-sequence frameworks~\cite{sutskever2014sequence} that directly encode online visual observations into the hidden state of recurrent neural units, which fail to capture structured spatial relationships~\cite{parisotto2018neural,tan2019learning}. Subsequent map-based methods introduce more explicit scene modeling, such as topological graphs~\cite{chaplot2020neural,deng2020evolving,chen2022think,wang2021structured,an2024etpnav} and top-down semantic maps~\cite{an2023bevbert,wang2023gridmm,chen2022weakly,hong2023learning,liu2023bird}. Although topological graphs are effective to capture abstract spatial relations, they lack 3D transformation equivariance, resulting in inconsistent spatial reasoning across viewpoints~\cite{li20223d,wang2023active}. Semantic maps, on the other hand, provide context-aware insights but struggle to model the 3D geometry necessary for precise spatial understanding~\cite{chen2022weakly,georgakis2022cross,hong2023learning}. Recent studies~\cite{wang2024lookahead,kwon2023renderable} have turned to implicit neural representations~\cite{mildenhall2020nerf} for map building, demonstrating impressive capabilities in capturing both 3D structures and semantics through continuous volumetric representations~\cite{zhi2021place,fu2022panoptic}. However, these representations typically employ dense and uniform volumetric sampling that covers the entire 3D volume, often failing to capture object boundaries and critical geometric structures~\cite{zhi2021place,gao2021dynamic} (see Fig.~\ref{fig_introduce}). They not only hinder accurate scene understanding, particularly in free and unoccupied spaces, but also lead to redundant representations and unnecessary computations~\cite{liu2021editing}. Additionally, existing methods are primarily trained in closed-vocabulary settings that lack the diversity to encompass the rich semantics and variations within VLN scenarios, thereby hampering their ability to generalize across unseen scenes~\cite{li2022envedit,liu2021vision,scheirer2012toward,fan2024evidential}.

To solve these problems, this work proposes a \textbf{3D Gaussian Map} that integrates geometric priors and open-set semantics, along with a corresponding navigation strategy to enhance sequential decision-making in VLN. The solution enables the agent to \textit{i)} construct 3D scene maps with geometric priors at each navigable point during navigation, \textit{ii)} integrate open-set semantics into the map, and \textit{iii)} incorporate the map into its decision-making process. In detail, \textbf{Egocentric Scene Map} (ESM, \S\ref{sec_EGM}) is introduced to represent the environment as a collection of differentiable 3D Gaussian primitives initialized from sparse pseudo-lidar point clouds. These primitives inherently preserve spatial structure and depth information, which serve as geometric priors that are essential for spatial awareness. Furthermore, \textbf{Open-Set Semantic Grouping} (OSG, \S\ref{sec_SSA}) operation is designed to bridge geometric and semantic understanding in ESM. OSG assigns an open-set semantic property to each Gaussian and groups them according to their object instance or stuff membership in the 3D scene. Based on this map, \textbf{Multi-Level Action Prediction} (MAP, \S\ref{sec_MAP}) strategy is crafted to facilitate navigation by aggregating information across scene, view and instance levels. The scene level leverages a global layout, the view level focuses on forward-facing cues, and the instance level enhances decision-making with precise semantic details.

Our method is evaluated on three public benchmarks: R2R~\cite{AndersonWTB0S0G18}, R4R~\cite{jain2019stay}, and REVERIE~\cite{qi2020reverie}. It achieves consistent improvements, with \textbf{2}\% gains in both SR and SPL on R2R, a \textbf{3}\% performance boost in SDTW on R4R, as well as \textbf{2.02}\% in RGS and \textbf{2.30}\% in RGSPL on REVERIE, all on the \textit{val unseen} splits (\S\ref{exp_comparison}). Comprehensive ablation studies validate the effectiveness of each component (\S\ref{exp_abalation}).


\section{Related Work}
\label{sec:related work}
\noindent\textbf{Vision-Language Navigation (VLN).} Early VLN approaches often rely on sequence-to-sequence models to establish connections between language and visual cues, encoding trajectory history within hidden states~\cite{fried2018speaker,AndersonWTB0S0G18,tan2019learning}. Subsequently, with advancements in transformer, VLN approaches have significantly improved cross-modal representations, which enable more precise alignment between visual scenes and linguistic instructions~\cite{hong2020language,wang2023dreamwalker}. Moreover, integrating imitation and reinforcement learning has proven beneficial in VLN, offering agents immediate guidance and facilitating long-term policy optimization for improved navigation outcomes~\cite{tan2019learning,hong2020language,wang2020active}. In addition, several studies are dedicated to grounding language by anchoring instructions through multimodal information fusion, thereby enhancing agents' ability to interpret and execute complex, multi-step directions~\cite{wang2019reinforced,zhong2024empowering}. Furthermore, to alleviate data scarcity and enhance the diversity of scenes in VLN, researchers have developed methods that emphasize environmental augmentation, instruction generation, and synthetic data creation. These approaches expand training resources and enhance the abilities of the agent to generalize across unseen and diverse scenarios~\cite{fan2024navigation,zhou2024migc,fan2024scene,kong2024controllable}.

Despite their contributions, most of them rely on 2D representations to encode environment information and predict actions. As a result, they struggle to capture the inherent complexity and spatial relationships of 3D scenes. In contrast, our method seamlessly integrates 3D geometry and semantics within a unified 3D Gaussian Map, enabling more informed decision-making based on its representations.

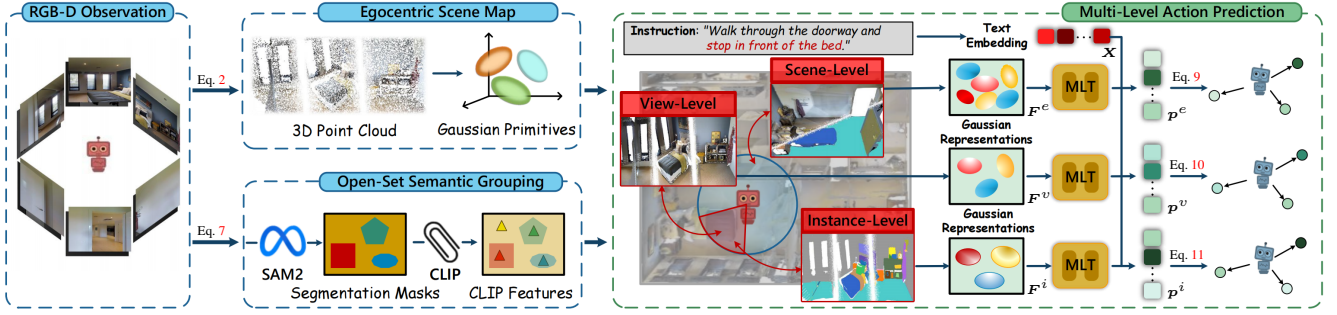
\begin{figure*}[t]
		\vspace{-10pt}
	\begin{center}
		\includegraphics[width=1\linewidth]{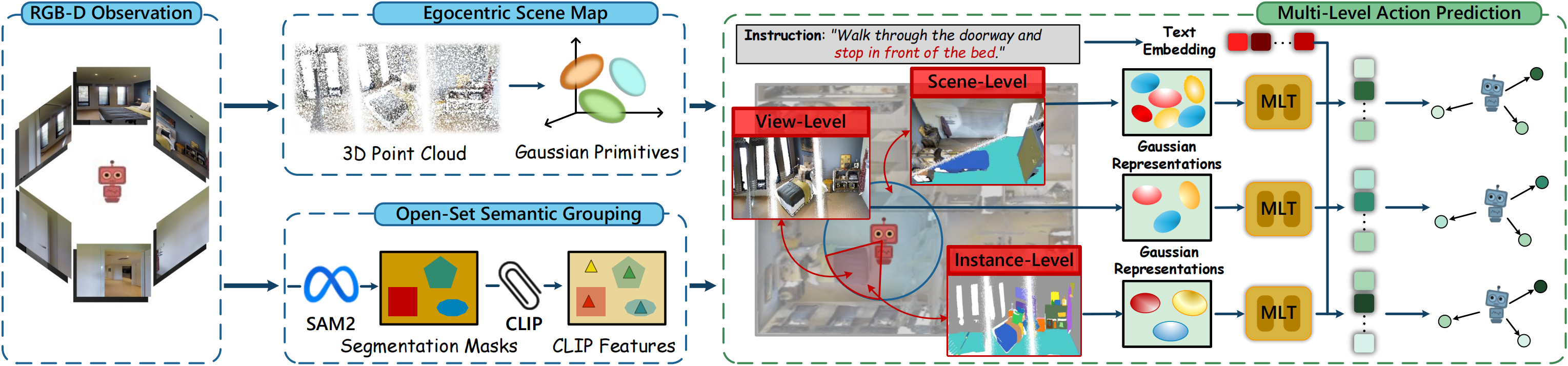}
        \put(-424,85){\tiny{Eq.~\ref{equ_lidarpoint2gs}}}
        \put(-424,28){\tiny{Eq.~\ref{eq_clip}}}
        \put(-84,95){\tiny{$\bm{X}$}}
        \put(-57,86){\tiny{Eq.~\ref{eq_scene}}}
        \put(-58,53){\tiny{Eq.~\ref{eq_view}}}
        \put(-58,19){\tiny{Eq.~\ref{eq_instance}}}
        \put(-112,73){\tiny{$\bm{F}^{e}$}}
        \put(-112,40){\tiny{$\bm{F}^{v}$}}
        \put(-112,6){\tiny{$\bm{F}^{\scriptscriptstyle{i}}$}}
        \put(-59,72){\tiny{$\bm{p}^{e}$}}
        \put(-59,37){\tiny{$\bm{p}^{v}$}}
        \put(-59,5){\tiny{$\bm{p}^{\scriptscriptstyle{i}}$}}


	\end{center}
	\vspace{-5pt}
	\captionsetup{font=small}
	\caption{\small{\textbf{Overview of our method.} At each node, our agent leverages egocentric RGB-D observations to generate pseudo-lidar point clouds, which are then used to initialize an \textit{Egocentric Scene Map} (\S\ref{sec_EGM}). Simultaneously, the observations are processed using \textit{Open-Set Semantic Grouping} (\S\ref{sec_SSA}) operation, which enriches the map with open-set semantic information. Based on this map, the agent employs the \textit{Multi-Level Action Prediction} (\S\ref{sec_MAP}) strategy to make informed navigation decisions. The scene level delivers a global layout, the view level emphasizes forward-facing features, and the instance level enhances decisions with fine-grained semantics. See \S\ref{sec:method} for more details.}}
	\label{fig_overview}
	\vspace{-1pt}
\end{figure*}

\noindent\textbf{Map Building.} In navigation tasks, map building is crucial for situational awareness and efficient path planning. Conventional approaches typically employ either topological or metric maps, each offering distinct advantages~\cite{quan2021holistic,an2023bevbert}. Metric maps provide precise spatial measurements, enabling direct distance calculations for path optimization~\cite{chaplot2020object,narasimhan2020seeing,georgakis2022cross}. In contrast, topological maps encode relational connections between key locations, supporting efficient node-to-node navigation in large-scale environments~\cite{chaplot2020neural,deng2020evolving,wang2021structured}. In addition, the improvement in SLAM and visual-language models has facilitated the emergence of semantic maps. Such maps integrate object- and scene-level information, allowing agents to interpret environments through contextual cues~\cite{chaplot2020object,wang2023gridmm,liu2023bird,wu2021vector}. Moreover, occupancy maps enhance navigation by modeling navigable and obstructed areas, dynamically updating the agent’s awareness of proximal free space and spatial layout in the scenes~\cite{georgakis2022cross,ramakrishnan2020occupancy,chen2022learning,liu2024volumetric}. Recent advancements in navigation have leveraged NeRF~\cite{mildenhall2020nerf} to enhance map representations. By encoding visual and geometric details into latent codes, NeRF enables view synthesis for richer scene understanding~\cite{devries2021unconstrained,kwon2023renderable,wang2024lookahead}.

However, the aforementioned methods generally do not explicitly encode geometric information, limiting their capacity to accurately capture scene-specific geometric structures and associated semantics~\cite{shim2023snerl}. In addition, volumetric representations often require dense and uniform sampling across 3D space. This results in a significant portion of samples lying in empty areas, leading to extra computational overhead~\cite{liu2024volumetric}. Unlike these methods, our 3D Gaussian Map encodes abundant geometric priors derived from RGB-D observations. Furthermore, due to the inherent sparsity and universal approximating ability of Gaussian mixtures~\cite{kerbl3Dgaussians}, this map captures fine-grained scene geometry and precise semantic information within the 3D environment.

\noindent\textbf{3D Scene Representations.} In VLN, 3D scene understanding is crucial as it allows the agent to perceive spatial structures, depth, and object relationships more realistically~\cite{yuan2023compositional}. Traditional 3D scene representations such as point clouds, meshes, or voxels can approximate spatial layouts, but they are computationally intensive and often fail to preserve detailed visual information~\cite{li2023voxformer,armeni20163d,dai2017scannet,jin2022deformation}. Subsequently, NeRF~\cite{mildenhall2020nerf} offers a breakthrough in 3D representation by rendering high-quality, continuous 3D scenes~\cite{zhi2021place,liu2021editing,gao2021dynamic,pumarola2021d}. Recently, 3D Gaussian Splatting (3DGS)~\cite{kerbl3Dgaussians}, renowned for its quality and speed, has been widely adopted across various domains to represent scenes by rendering radiance fields with multiple 3D Gaussians~\cite{charatan2024pixelsplat,chen2024text,liang2024luciddreamer,keetha2024splatam,jiang2024hifi4g,zhou2024feature,shi2024language,chen2024survey,guo2025multi}. 

However, these methods primarily focus on incrementally building a single global map, which is mainly used for scene synthesis and editing. Moreover, in the original 3DGS~\cite{kerbl3Dgaussians}, each Gaussian is parameterized by its position, scale, rotation, opacity, and color. To capture task-specific information, several studies have adapted 3DGS by incorporating additional attributes such as linguistic, semantic, and spatio-temporal properties~\cite{xu20244k4d,ye2025gaussian,zhou2024hugs}. In contrast, our approach is designed to support decision-making in VLN by constructing multiple egocentric maps during navigation. Additionally, it leverages SAM2~\cite{ravi2024sam} and CLIP~\cite{radford2021learning} for structured semantic alignment, thereby enhancing the capability of agents for 3D spatial awareness.


\section{Method}
\label{sec:method}
\noindent\textbf{Problem Formulation.} In VLN, an agent traverses a 3D environment guided by natural language instructions $\mathcal{X}$ to reach a target location~\cite{AndersonWTB0S0G18} or identify an object~\cite{qi2020reverie}. The 3D environment is typically modeled as a discretized navigable graph~\cite{chang2017matterport3d}, consisting of a set of nodes as viewpoints and connectivity edges for movement. At each navigation step $t$, the agent receives a 360-degree panoramic observation comprising RGB images $\mathcal{I}_t\!=\!\{I_{t,k}\}_{k=1}^K$ and associated depth images $\mathcal{D}_t\!=\!\{D_{t,k}\}_{k=1}^{K}$, where $I_{t,k}\!\in\!\mathbb{R}^{H\times W\times3}$ and $_{\!}$$D_{t,k}\!\in\!\mathbb{R}^{H\times W}$ $_{\!}$denote the images captured in the $k$-th direction. Built upon this, the agent is required to learn a navigation policy that predicts the next step action $a_t\in\mathcal{A}_t$. The action space $\mathcal{A}_t$ comprises $N_t$ neighboring nodes $\mathcal{V}_t=\{V_{t,n}\}_{n=1}^{N_t}$, other observed nodes $\mathcal{V}_t^*$ (through backtrack~\cite{wang2021structured,chen2022think}), and a [STOP] option.

\noindent\textbf{Overview.} At each node, the agent initializes 3D Gaussians from multi-view RGB-D observations to build \textit{Egocentric Scene Map} (ESM, \S\ref{sec_EGM}), while simultaneously enhancing these Gaussians through \textit{Open-Set Semantic Grouping} (OSG, \S\ref{sec_SSA}) operation. Based on this map, the agent performs \textit{Multi-Level Action Prediction} (MAP, \S\ref{sec_MAP}) strategy, using multi-level cues for decision-making (see Fig.~\ref{fig_overview}).

\subsection{Egocentric Scene Map (ESM)}\label{sec_EGM}
ESM models the spatial structure of scenes using differentiable 3D Gaussians, initialized from sparse pseudo-lidar point clouds derived from multi-view RGB-D observations. In addition to inheriting geometric priors from the point clouds, ESM leverages the universal approximation capability of Gaussian mixtures~\cite{kerbl3Dgaussians} to capture fine-grained spatial structures, thereby providing a robust foundation for semantic enrichment and decision-making.

\noindent\textbf{Initialization.} At time step $t$, multi-view RGB-D observations $\{\mathcal{I}_t,\mathcal{D}_t\}$ are back-projected into the pseudo-lidar point cloud $\mathcal{P}_t$. Each pixel $(u, v)$ in the image $I_{t,k}$ will be transformed into 3D coordinates $(x, y, z)$ as follows:
\begin{equation}\small
\begin{aligned}
z=D_{t,k}(u,v),~~x=\frac{(u-c^u)z}{f^x},~~y=\frac{(v-c^v)z}{f^y},
\end{aligned}
\label{equ_lidarpoint}
\end{equation}
where $D_{t,k}(u, v)$ represents the depth of the pixel in camera coordinates, $(c^u, c^v)$ denotes the camera center, and $f^x$ and $f^y$ are the horizontal and vertical focal length of the camera. After the transformation between camera and world coordinate system, this point cloud serves as a geometric prior for initializing the 3D Gaussian primitives $\mathcal{G}_t\!=\!\{\bm{g}_{t,i}\}_i^{|\mathcal{P}_t|}$. The 2D-to-3D mapping process $\mathcal{M}^{{\rm2D}\rightarrow \rm{3D}}$ is defined as:
\begin{equation}\small
\begin{aligned}
\mathcal{G}_t=\mathcal{M}^{\rm{2D}\rightarrow \rm{3D}}(\mathcal{I}_t,\mathcal{D}_t).
\end{aligned}
\label{equ_lidarpoint2gs}
\end{equation}
In addition to the geometric prior (\textit{i.e.}, the position $\bm{\mu}_i\!=\!(x_i,y_i,z_i)\!\in\!\mathbb{R}^3$ for the centroid), each Gaussian primitive is also initialized with a set of additional parameters, \textit{i.e.}, covariance matrix $\bm{\Sigma}_i\!\in\!\mathbb{R}^{3 \times 3}$, opacity $\alpha_i\!\in\![0,1]$, and color vector $\bm{c}_i\!\in\!\mathbb{R}^3$. $t$ is omitted for simplicity. Specifically, $\bm{\Sigma}_i\!=\!\bm{R}\bm{S}\bm{S}^\top \bm{R}^\top$ encodes scale and orientation, where the rotation matrix $\bm{R}$ and the scale matrix $\bm{S}$ are stored as a 3D vector $\bm{s}_i\!\in\!\mathbb{R}^3$ and a quaternion $\bm{r}_i\!\in\!\mathbb{R}^4$, respectively, for independent optimization. Moreover, $\alpha_i$ adjusts transparency for $\alpha$-blending of anisotropic splats, while $\bm{c}_i$ enables view-dependent appearance with spherical harmonics.

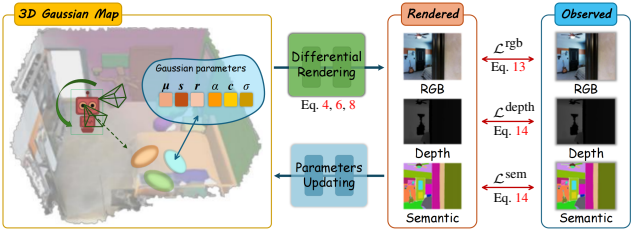
\begin{figure}[t]
	\begin{center}
		\includegraphics[width=1\linewidth]{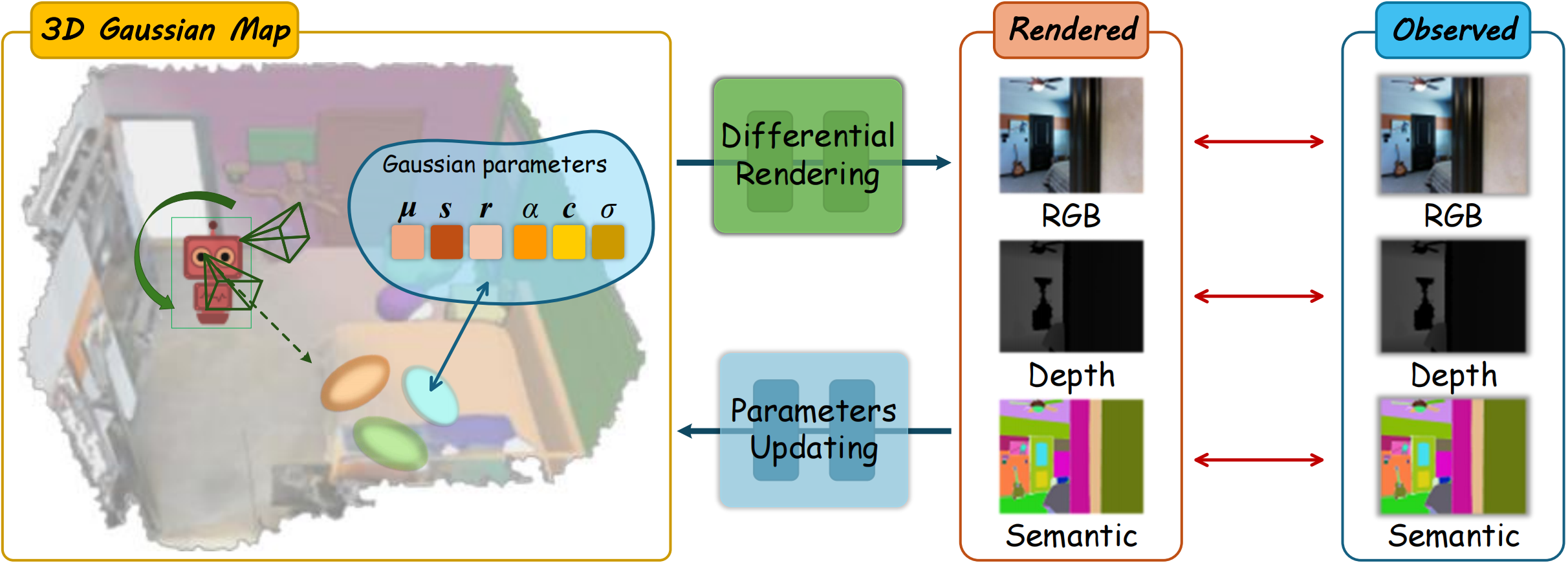}
        \put(-124.5,45){\tiny{Eq.~\ref{equ_RGB},~\ref{eq_depth},~\ref{eq_semantic}}}
        \put(-53,59){\tiny{Eq.~\ref{eq_lrgb}}}
        \put(-52,35){\tiny{Eq.~\ref{eq_lds}}}
        \put(-52,10){\tiny{Eq.~\ref{eq_lds}}}
        \put(-52.5,66){\tiny{$\mathcal{L}^{\text{rgb}}$}}
        \put(-52.5,42){\tiny{$\mathcal{L}^{\text{depth}}$}}
        \put(-52.5,17){\tiny{$\mathcal{L}^{\text{sem}}$}}
        
	\end{center}
	\vspace{-7pt}
	\captionsetup{font=small}
	\caption{\small{\textbf{3D Gaussian Map Optimization}. Gaussian parameters (position $\bm{\mu}$, scale $\bm{s}$, rotation $\bm{r}$, opacity ${\alpha}$, color $\bm{c}$, and semantic ${\sigma}$) are optimized through the differential rendering process, where the parameters are updated using RGB, depth, and semantic losses ($\mathcal{L}^{\text{rgb}}$, $\mathcal{L}^{\text{depth}}$, $\mathcal{L}^{\text{sem}}$). See \S\ref{sec:method} for more details.}}
    	\label{fig_coarsetofine}
	\vspace{-1pt}
\end{figure}

\noindent\textbf{Differentiable Construction.} After initializing Gaussian primitives $\mathcal{G}_t$, a tile-based renderer $\mathcal{M}^{{\rm3D}\rightarrow \rm{2D}}$ rasterizes these primitives to synthesize corresponding 2D observation $\{\hat{I}_t,\hat{D}_t\}$ of the scene from a specific camera pose:
\begin{equation}\small
\begin{aligned}
\hat{I}_t,\hat{D}_t=\mathcal{M}^{\rm{3D}\rightarrow \rm{2D}}(\mathcal{G}_t).
\end{aligned}
\end{equation}
Each pixel value $\hat{I}_t(u,v)$ of the rendered 2D observation is derived by blending depth-ordered Gaussians~\cite{kerbl3Dgaussians}:
\begin{equation}\small
\begin{aligned}
\hat{I}_t(u,v) = \sum\nolimits_{i} \bm{c}_i \alpha_{i}' \prod\nolimits_{j=1}^{i-1} (1 - \alpha_{j}')\in\mathbb{R}^{3},
\end{aligned}
\label{equ_RGB}
\end{equation}
where $i$ indicates the depth ordering of Gaussians overlapping at pixel $(u,v)$. $\alpha_{i}'$ is calculated based on $\alpha_{i}$ and an exponential decay factor related to the pixel offset:
\begin{equation}\small
\begin{aligned}
\alpha_{i}' = \alpha_i \cdot \exp \big( -\frac{1}{2} (\bm{x}' - \bm{\mu}'_i)^\top \bm{\Sigma}'^{-1}_i (\bm{x}' - \bm{\mu}'_i) \big)\in\mathbb{R}^+,
\end{aligned}
\end{equation}
where $\bm{x}'=(u,v)$ and $\bm{\mu}'_i\!\in\!\mathbb{R}^2$ represents the coordinates on the transformed 2D plane. $\bm{\Sigma}'_i$ denotes the splatted 2D version of $\bm{\Sigma}_i$. 
Similarly, an analogous differentiable rendering process is applied to compute the depth $\hat{D}_t(u,v)$ at each pixel of the specific camera pose:
\begin{equation}\small
\begin{aligned}
\hat{D}_t(u,v) = \sum\nolimits_{i} z_{i} \alpha_{i}' \prod\nolimits_{j=1}^{i-1} (1 - \alpha_{j}') \in\mathbb{R}^+,
\end{aligned}
\label{eq_depth}
\end{equation}
where $z_{i}$ is the distance to the center of the Gaussian $\bm{g}_i$ along the camera ray. The differentiable rendering process enables gradients from pixel-level loss functions to back-propagate through the Gaussian parameters. As a result, by iteratively minimizing the error between rendered and observed RGB-D images, ESM is progressively constructed.

\subsection{Open-Set Semantic Grouping (OSG)}\label{sec_SSA}
While ESM inherits informative geometric priors from the pseudo-lidar point clouds, it lacks semantic information, which is essential for comprehending complex spatial relationships and adapting to diverse VLN scenarios. To bridge this gap, we introduce OSG operation, enriching ESM with open-set semantics by associating each Gaussian primitive with semantic properties derived from visual observations.

\noindent\textbf{Open-Set $_{\!}$Semantic $_{\!}$Encoding.} $_{\!}$At $_{\!}$step $_{\!}$$t$, $_{\!}$SAM2~\cite{ravi2024sam} is used to automatically generate 2D masks $\{\bm{m}_1, \bm{m}_2, \dots, \bm{m}_K\}$ in \textit{everything mode} for the panoramic observation $\mathcal{I}_t=\{I_{t,k}\}_{k=1}^K$. Each $\bm{m}_k \in \mathbb{R}^{H_k \times W_k \times 3}$ captures a spatially coherent region within the scene. Semantic embeddings for each region are derived via CLIP~\cite{radford2021learning} expressed as: 
\begin{equation}\small
\begin{aligned}
\bm{F}^s_k=\mathcal{F}^\text{CLIP}(\bm{m}_k)\in\mathbb{R}^{512}.
\end{aligned}
\label{eq_clip}
\end{equation}
In addition, storing full language embeddings incurs significant memory overhead, even though a single scene typically occupies only a limited portion of the CLIP feature space. To address this, global average pooling is applied to $\bm{F}^s_k$, producing a more compact semantic encoding $F^s_k \in \mathbb{R}$. 

\noindent\textbf{Semantic Grouping via Rendering.} With the compact semantic encoding, we integrate these semantics into ESM via a rendering process similar to the color and depth optimization. Specifically, an additional semantic parameter $\sigma \in \mathbb{R}$ is introduced for each Gaussian $\bm{g}_i$. Each $\sigma$ is randomly initialized and refined through the same rendering process. Like Eq.~\ref{equ_RGB}, the semantic representation $\hat{F^s}$ for each pixel in 2D image space is obtained by aggregating $\sigma_{i}$ of depth-ordered Gaussians, weighted by opacity $\alpha_{i}'$:
\begin{equation}\small
\begin{aligned}
\hat{F^s} = \sum\nolimits_{i} \sigma_{i} \alpha_{i}' \prod\nolimits_{j=1}^{i-1} (1 - \alpha_{j}')  \in \mathbb{R}.
\end{aligned}
\label{eq_semantic}
\end{equation}
Instead of relying on manual 3D annotations, $\hat{F^s}$ is optimized in parallel with target CLIP embeddings during the differentiable construction of ESM. This process establishes semantic associations between Gaussians and harmonizes open-set semantics from OSG with geometric priors in ESM, resulting in a unified 3D Gaussian Map.

\subsection{Multi-Level Action Prediction (MAP)}\label{sec_MAP}
The 3D Gaussian Map $\mathcal{G}$, constructed by integrating ESM and OSG, consists of Gaussians $\bm{g}_i$ parameterized by $\{\bm{\mu}_i,\bm{s}_i,\bm{r}_i,\alpha_i,\bm{c}_i,\sigma_i\}$. For ease of notation, we reuse $\bm{g}_i \in \mathbb{R}^{7}$ to denote the Gaussian representation of this map, which is a concatenated vector of the mean $\bm{\mu}_i \in \mathbb{R}^{3}$, color $\bm{c}_i \in \mathbb{R}^{3}$, and semantics $\sigma_i \in \mathbb{R}$. Based on $\bm{g}$, we design MAP strategy to predict action probabilities by aggregating spatial-semantic cues from candidate waypoints $\mathcal{V}$, guided by the $L$-word instruction embedding $\bm{X}\!\in\!\mathbb{R}^{L\times 768}$. This strategy is structured across three levels: scene, view, and instance. $t$ is omitted for simplicity.

\noindent\textbf{Scene Level.} This level aggregates information from the entire 3D Gaussian Map $\mathcal{G}$ to provide a global understanding of the environment. The scene feature $\bm{F}^{e}$ is computed using global average pooling over all Gaussian representations $\bm{g}_i$ in $\mathcal{G}$, providing a holistic representation of the scene. The scene-level score $\bm{p}^e$ is derived by applying multi-layer transformers with feed-forward layers (MLT) $\mathcal{F}^\text{MLT}$~\cite{chen2022think}, offering spatial guidance to the agent. This is formulated as follows (where $[\cdot,\cdot]$ denotes concatenation):
\begin{equation}\small
\begin{aligned}
\bm{p}^{e} = \text{Softmax}(\mathcal{F}^\text{MLT}([\bm{F}^{e},\bm{X}])) \in [0,1]^{|\mathcal{V}|},
\end{aligned}
\label{eq_scene}
\end{equation}
where $|\mathcal{V}|$ indicates the number of candidate points.

\noindent\textbf{View Level.} This level restricts the agent’s attention to Gaussians within its current observation, exploiting spatial information aligned with the movement direction to support decision-making. By aggregating the selected representations $\bm{g}_i$, the view feature $\bm{F}^{v}$ is generated. This feature is then transformed by $\mathcal{F}^\text{MLT}$ to yield the view-level score $\bm{p}^v$:
\begin{equation}\small
\begin{aligned}
\bm{p}^{v} = \text{Softmax}(\mathcal{F}^\text{MLT}([\bm{F}^{v},\bm{X}])) \in [0,1]^{|\mathcal{V}|}.
\end{aligned}
\label{eq_view}
\end{equation}

\noindent\textbf{Instance Level.} This level further focuses on individual instances within the current observation, capturing fine-grained details to enable precise and context-aware trajectory adjustments. For each identified instance, features are derived by aggregating its associated Gaussian representations $\bm{g}_i$. These features are then stacked into a combined representation $\bm{F}^{i}$, followed by $\mathcal{F}^\text{MLT}$ to generate the instance-level score $\bm{p}^{i}$:
\begin{equation}\small
\begin{aligned}
\bm{p}^{i} = \text{Softmax}(\mathcal{F}^\text{MLT}([\bm{F}^{i},\bm{X}])) \in [0,1]^{|\mathcal{V}|}.
\end{aligned}
\label{eq_instance}
\end{equation}

\noindent\textbf{Multi-Level Scores.} To utilize multi-level information for decision-making, scene-, view-, and instance-level scores are integrated into candidate node probabilities $\bm{p}^{c}$, which are aligned with the action space $\mathcal{A}$:
\begin{equation}\small
\begin{aligned}
\bm{p}^{c}=\mathcal{N}(\bm{p}^{e},\mathcal{V})+\mathcal{N}(\bm{p}^{v},\mathcal{V})+\mathcal{N}(\bm{p}^{i},\mathcal{V})  \in [0,1]^{|\mathcal{V}|},
\end{aligned}
\label{eq_score}
\end{equation}
where $\mathcal{N}$ denotes the mapping of scores to nearby candidate nodes $\mathcal{V}$ using a nearest neighbor search. In this manner, MAP refines the agent’s spatial-semantic understanding across multiple scales, ranging from global contextual awareness to fine-grained navigation cues. This process iterates until the agent successfully reaches the destination.

\begin{table*}[t]
\centering
	\vspace{-10pt}
        \resizebox{1\textwidth}{!}{
		\setlength\tabcolsep{2.5pt}
		\renewcommand\arraystretch{1.0}
\begin{tabular}{c||cccccc|cccccc|cccccc}
\hline \thickhline
\rowcolor{mygray}
~ &  \multicolumn{18}{c}{REVERIE} \\
\cline{2-19}
\rowcolor{mygray}
~ &  \multicolumn{6}{c|}{\textit{val} \textit{seen}} & \multicolumn{6}{c|}{\textit{val} \textit{unseen}} & \multicolumn{6}{c}{\textit{test} \textit{unseen}} \\
\cline{2-19}
\rowcolor{mygray}
\multirow{-3}{*}{Models} &\small{TL$\downarrow$} &\small{OSR$\uparrow$} &\small{SR$\uparrow$} &\small{SPL$\uparrow$} &\small{RGS$\uparrow$} &\small{RGSPL$\uparrow$} &\small{TL$\downarrow$} &\small{OSR$\uparrow$} &\small{SR$\uparrow$} &\small{SPL$\uparrow$} &\small{RGS$\uparrow$} &\small{RGSPL$\uparrow$} &\small{TL$\downarrow$} &\small{OSR$\uparrow$} &\small{SR$\uparrow$} &\small{SPL$\uparrow$} &\small{RGS$\uparrow$} &\small{RGSPL$\uparrow$}\\
\hline
\hline
RCM~\cite{wang2019reinforced}    &10.70 &29.44 &23.33 &21.82 &16.23 &15.36    &11.98 &14.23 &9.29  &6.97  &4.89  &3.89      &10.60 &11.68 &7.84  &6.67  &3.67  &3.14  \\
FAST-M~\cite{qi2020reverie}  &16.35 &55.17 &50.53 &45.50 &31.97 &29.66    &45.28 &28.20 &14.40 &7.19  &7.84  &4.67      &39.05 &30.63 &19.88 &11.61 &11.28 &6.08  \\
SIA~\cite{lin2021scene}          &13.61 &65.85 &61.91 &57.08 &45.96 &42.65    &41.53 &44.67 &31.53 &16.28 &22.41 &11.56     &48.61 &44.56 &30.80 &14.85 &19.02 &9.20  \\
RecBERT~\cite{hong2021vln}        &13.44 &53.90 &51.79 &47.96 &38.23 &35.61    &16.78 &35.02 &30.67 &24.90 &18.77 &15.27     &15.86 &32.91 &29.61 &23.99 &16.50 &13.51 \\
Airbert~\cite{guhur2021airbert}       &15.16 &48.98 &47.01 &42.34 &32.75 &30.01    &18.71 &34.51 &27.89 &21.88 &18.23 &14.18     &17.91 &34.20 &30.28 &23.61 &16.83 &13.28 \\
HAMT~\cite{chen2021history}            &12.79 &47.65 &43.29 &40.19 &27.20 &25.18    &14.08 &36.84 &32.95 &30.20 &18.92 &17.28     &13.62 &33.41 &30.40 &26.67 &14.88 &13.08 \\
HOP~\cite{qiao2022hop}           &13.80 &54.88 &53.76 &47.19 &38.65 &33.85    &16.46 &36.24 &31.78 &26.11 &18.85 &15.73     &16.38 &33.06 &30.17 &24.34 &17.69 &14.34 \\
DUET~\cite{chen2022think}           &13.86 &73.86 &71.75 &63.94 &57.41 &51.14    &22.11 &51.07 &46.98 &33.73 &32.15 &23.03     &21.30 &56.91 &52.51 &36.06 &31.88 &22.06 \\
GridMM~\cite{wang2023gridmm}        &$-$   &$-$   &$-$   &$-$   &$-$   &$-$    &23.20 &57.48 &51.37 &36.47 &34.57 &24.56 &19.97 &59.55 &53.13 &36.60 &34.87 &23.45 \\
LANA~\cite{wang2023lana}         &15.91 &74.28 &71.94 &62.77 &59.02 &50.34    &23.18 &52.97 &48.31 &33.86 &32.86 &22.77     &18.83 &57.20 &51.72 &36.45 &32.95 &22.85\\
BEVBert~\cite{an2023bevbert}        &$-$   &76.18   &73.72   &65.32   &57.70   &51.73    &$-$   &56.40 &51.78 &36.37 &34.71 &24.44 &$-$   &57.26 &52.81 &36.41 &32.06 &22.09 \\
\hline
\textbf{Ours}  & 13.94 & \textbf{77.21} & \textbf{74.96} & \textbf{66.50} & \textbf{59.41} & \textbf{52.70} & 22.22 & \textbf{58.81} & \textbf{53.59} & \textbf{37.67} & \textbf{36.73} & \textbf{26.74} & 20.05 & 56.93 & \textbf{52.93} & \textbf{36.93} & \textbf{35.65} & \textbf{25.76} \\
\hline
\end{tabular}
}
	\vspace*{-5pt}
\captionsetup{font=small}
	\caption{\small{\textbf{Quantitative results} on REVERIE~\cite{qi2020reverie}. `$-$': unavailable statistics. See \S\ref{exp_comparison} for more details.}}
    \label{table_REVERIE}
\vspace*{-10pt}
\end{table*}

\subsection{Loss Function for Gaussian Rendering}
\noindent\textbf{3D Gaussian Map Losses.} A combination of $\mathcal{L}^1$ and Structural Similarity~\cite{wang2004image} (SSIM) loss is used to optimize the rendered color $\hat{I}$ with respect to the ground truth $I$:
\begin{equation}\small
\begin{aligned}
    \mathcal{L}^{\text{rgb}} = (1 - \lambda^{\text{SSIM}}) \big\| \hat{I} - I \big\|_1 + \lambda^{\text{SSIM}} \cdot \text{SSIM}(\hat{I}, I).
\end{aligned}
\label{eq_lrgb}
\end{equation}
The depth map $\hat{D}$ is supervised by $\mathcal{L}^1$ against the ground truth depth $D$, while the semantic feature $\hat{{F}^s}$ is aligned with the target CLIP embedding ${F}^s$:
\begin{equation}\small
\mathcal{L}^{\text{depth}} = \big\| \hat{D} - D \big\|_1, \quad \mathcal{L}^{\text{sem}} = \big\| \hat{{F}^s} - {F}^s \big\|_1.
\label{eq_lds}
\end{equation}
These losses iteratively refine the 3D Gaussian Map through the differentiable rendering process, progressively integrating geometric priors and open-set semantic information.

\subsection{Implementation Details}
\noindent\textbf{Topological Memory.} Following prior works~\cite{chen2022think, liu2024volumetric}, to support long-time and context-aware navigation, we adopt a topological memory mechanism that dynamically updates as the agent explores the environment. This memory stores both visited and navigable nodes, along with information derived from the 2D panorama and the 3D Gaussian Map. These elements collectively form a graph-like structure, where edges represent possible transitions. The multi-level navigation scores, combined with the traditional 2D action score~\cite{chen2022think}, jointly evaluate and rank these transitions. During navigation, the memory allows the agent to revisit previously explored regions or evaluate alternative paths, thereby reducing uncertainty in complex layouts. By leveraging the stored 3D Gaussian Map, which provides spatially coherent geometric and semantic information, the agent is able to make informed decisions (see more details in \textbf{Appendix}).

\noindent\textbf{3D Gaussian Map.} To ensure efficiency and sparse sampling, the RGB-D observations are resized to $224\times224$, and the 3D Gaussian Map is constructed at this resolution. Offline pretraining is conducted on a single NVIDIA RTX 4090 GPU for 15 iterations (see more details in \textbf{Appendix}).

\noindent\textbf{Network Pretraining.} For R2R~\cite{AndersonWTB0S0G18} and R4R~\cite{jain2019stay}, Masked Language Modeling (MLM)~\cite{kenton2019bert,chen2021history} and Single-step Action Prediction (SAP)~\cite{chen2021history,hong2021vln} are adopted as auxiliary objectives during pretraining. Moreover, for REVERIE~\cite{qi2020reverie}, we additionally introduce Object Grounding (OG)~\cite{chen2022think,lin2021scene} to enhance object-level reasoning. Pretraining is conducted with a batch size of 64 over 100k iterations, using the Adam optimizer~\cite{KingmaB14adam} with a learning rate of 1e-4.

\noindent\textbf{Network Finetuning.} Following classical paradigm~\cite{chen2022think}, the pretrained model is finetuned using DAgger~\cite{RossGB11}. For REVERIE~\cite{qi2020reverie}, an OG loss term, weighted at 0.20, is incorporated to balance object grounding and navigation tasks. Finetuning is performed over 25k iterations with a batch size of 8 and a learning rate of 1e-5. Optimal iterations are determined based on peak performance on \textit{val unseen} splits.

\noindent\textbf{Testing.} At each waypoint, our agent constructs the 3D Gaussian Map using multi-view RGB-D observations and applies MAP strategy to assist in its decision-making process. This process concludes when the agent either reaches the target or selects [STOP]. In addition, during navigation, constructing the 3D Gaussian Map at each time step takes approximately \textbf{0.07} seconds, ensuring compatibility with real-time robotic execution (see more details in \textbf{Appendix}).

\begin{table}[t]
\centering
        \resizebox{0.49\textwidth}{!}{
		\setlength\tabcolsep{3pt}
		\renewcommand\arraystretch{1.0}
\begin{tabular}{c||cccc|cccc}
\hline \thickhline
\rowcolor{mygray}
~ &  \multicolumn{8}{c}{R2R} \\
\cline{2-9}
\rowcolor{mygray}
~ & \multicolumn{4}{c|}{\textit{val} \textit{unseen}} & \multicolumn{4}{c}{\textit{test} \textit{unseen}} \\
\cline{2-9}
\rowcolor{mygray}
\multirow{-3}{*}{Models} &TL$\downarrow$ &NE$\downarrow$ &SR$\uparrow$ &SPL$\uparrow$ &TL$\downarrow$ &NE$\downarrow$ &SR$\uparrow$ &SPL$\uparrow$\\
\hline
\hline
Seq2Seq~\cite{AndersonWTB0S0G18}  &8.39    &7.81 &22 &$-$    &8.13  &7.85 &20 &18 \\
SF~\cite{fried2018speaker}        &$-$     &6.62 &35 &$-$    &14.82 &6.62 &35 &28 \\
EnvDrop~\cite{tan2019learning}            &10.70   &5.22 &52 &48     &11.66 &5.23 &51 &47 \\
AuxRN~\cite{zhu2020vision}        &$-$     &5.28 &55 &50     &$-$   &5.15 &55 &51 \\
Active~\cite{wang2020active}   &20.60   &4.36 &58 &40     &21.60 &4.33 &60 &41\\
RecBERT~\cite{hong2021vln}         &12.01   &3.93 &63 &57     &12.35 &4.09 &63 &57 \\
HAMT~\cite{chen2021history}             &11.46   &2.29 &66 &61     &12.27 &3.93 &65 &60 \\
SOAT~\cite{moudgil2021soat}       &12.15   &4.28 &59 &53     &12.26 &4.49 &58 &53 \\
SSM~\cite{wang2021structured}     &20.7    &4.32 &62 &45     &20.4  &4.57 &61 &46 \\
CCC~\cite{wang2022counterfactual} &$-$     &5.20 &50 &46     &$-$   &5.30 &51 &48 \\
HOP~\cite{qiao2022hop}            &12.27   &3.80 &64 &57     &12.68 &3.83 &64 &59 \\
DUET~\cite{chen2022think}            &13.94   &3.31 &72 &60     &14.73 &3.65 &69 &59 \\
LANA~\cite{wang2023lana}          &12.0 &$-$ &68 &62      &12.6  &$-$ &65 &60\\
TD-STP~\cite{zhao2022target}    &$-$  &3.22 &70 &63      &$-$ &3.73 &67 &61\\
BSG~\cite{liu2023bird}            &14.90   &2.89 &74 &62    &14.86 &3.19 &73 &62 \\
BEVBert~\cite{an2023bevbert}      &14.55   &2.81 &75 &64     &$-$   &3.13 &73 &62 \\
\hline
\textbf{Ours}             & 14.83 & 2.43 & \textbf{77} & \textbf{66} & 14.58 & 3.17 & \textbf{75} & \textbf{65} \\
\hline
\end{tabular}
}
	\vspace*{-5pt}
\captionsetup{font=small}
	\caption{\small{\textbf{Quantitative results} on R2R~\cite{AndersonWTB0S0G18} (\S\ref{exp_comparison}).}}
    \label{table_R2R}
\vspace*{-10pt}
\end{table}

\section{Experiment}
\label{sec:exp}
\subsection{Experimental Setup}
\noindent\textbf{Datasets.} We evaluate our method on three benchmark datasets: R2R~\cite{AndersonWTB0S0G18}, R4R~\cite{jain2019stay}, and REVERIE~\cite{qi2020reverie}. R2R contains 7,189 trajectories, each paired with three natural language instructions, split into \textit{train}, \textit{val seen}, \textit{val unseen}, and \textit{test unseen} sets spanning 61, 56, 11, and 18 scenes, respectively. R4R extends R2R by concatenating adjacent trajectories into longer instructions. REVERIE requires the agent to locate targets from high-level instructions and select the correct bounding box upon reaching the goal.

\noindent\textbf{Evaluation Metrics.} The performance is evaluated using Trajectory Length (TL), Navigation Error (NE), Success Rate (SR), and Success-weighted Path Length (SPL), following~\cite{liu2021vision}. TL and NE assess distance efficiency, whereas SR and SPL indicate task success. For R4R, additional metrics include Coverage Length Score (CLS), Normalized Dynamic Time Warping (NDTW) for path fidelity, and Success-weighted Dynamic Time Warping (SDTW) for balancing accuracy with SR. On REVERIE, Remote Grounding Success (RGS) and its SPL-weighted variant (RGSPL) evaluate object grounding accuracy. Higher scores indicate better performance for all metrics except TL and NE.

\begin{figure*}[t]
	\begin{center}
		\includegraphics[width=0.98\linewidth]{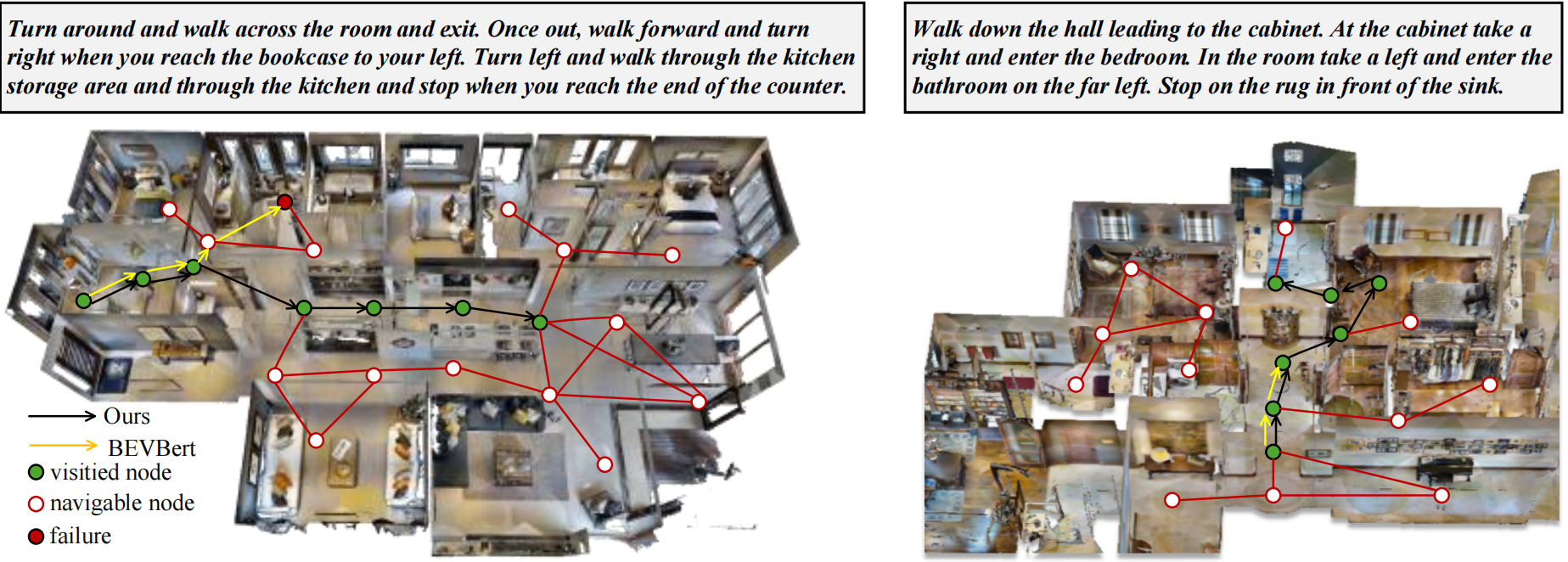}
        \put(-361,-8){{(a)}} 
        \put(-107,-8){{(b)}}
        
	\end{center}
	\vspace{-15pt}
	\captionsetup{font=small}
	\caption{\small{\textbf{Qualitative results} on R2R~\cite{AndersonWTB0S0G18} \textit{val unseen} split. (a) Our agent successfully navigates through multiple rooms and recognizes key landmarks, such as the \textit{``bookcase"} and \textit{``kitchen storage area"}, demonstrating the effectiveness of our 3D Gaussian Map in integrating geometric and semantic information. In contrast, BEVBert~\cite{an2023bevbert} deviates by selecting an incorrect room soon after leaving the \textit{``bedroom"}. (b) Our agent precisely identifies and localizes the \textit{``bathroom"} and \textit{``rug"}, while BEVBert~\cite{an2023bevbert} stops in the wrong place since critical landmarks cannot be identified, highlighting the fine-grained semantic awareness of our method. See \S\ref{exp_comparison} for more details.}}
	\label{exp_vis_ab}
	\vspace{-4pt}
\end{figure*}

\subsection{Comparison to State-of-the-Arts}\label{exp_comparison}
\noindent\textbf{Performance on REVERIE~\cite{qi2020reverie}.} Table~\ref{table_REVERIE} lists the overall performance on REVERIE. This dataset challenges the agent to locate specific objects at target location based on high-level instructions that only describe abstract goals. Our method outperforms BEVBert~\cite{an2023bevbert} by \textbf{2.02}\% in RGS and \textbf{2.30}\% in RGSPL on the \textit{val unseen} split, underscoring its effectiveness in accurate object grounding for VLN.

\noindent\textbf{Performance on R2R~\cite{AndersonWTB0S0G18}.} Table~\ref{table_R2R} compares our approach with recent methods on R2R. Our agent achieves consistent improvements across all splits, which outperforms BEVBert~\cite{an2023bevbert} by \textbf{2}\% in both SR and SPL on the \textit{val unseen} split. These results clearly underscore the effectiveness of our 3D Gaussian Map in advancing VLN performance.

\begin{table}[t]
\centering
        \resizebox{0.49\textwidth}{!}{
		\setlength\tabcolsep{7pt}
		\renewcommand\arraystretch{1.0}
\begin{tabular}{c||ccccc}
\hline \thickhline
\rowcolor{mygray}
~ & \multicolumn{5}{c}{R4R \textit{val} \textit{unseen}}  \\
\cline{2-6}
\rowcolor{mygray}
\multirow{-2}{*}{Models}    &NE$\downarrow$ &SR$\uparrow$ &CLS$\uparrow$ &nDTW$\uparrow$ &SDTW$\uparrow$ \\
\hline
\hline
SF~\cite{AndersonWTB0S0G18}        &8.47    &24   &30   &$-$   &$-$  \\
RCM~\cite{wang2019reinforced}     &$-$     &29   &35   &30    &13   \\
EGP~\cite{deng2020evolving}       &8.00    &30   &44   &37    &18   \\
SSM~\cite{wang2021structured}     &8.27    &32   &53   &39    &19   \\
RelGraph~\cite{hong2020language}  &7.43    &36   &41   &47    &34   \\
RecBERT~\cite{hong2021vln}         &6.67    &44   &51   &45    &30   \\
HAMT~\cite{chen2021history}        &6.09    &45   &58   &50    &32   \\
\hline
\textbf{Ours}  & \textbf{6.05} & \textbf{47} & \textbf{60} & \textbf{52} & \textbf{35} \\
\hline
\end{tabular}
}
	\vspace*{-5pt}
\captionsetup{font=small}
	\caption{\small{\textbf{Quantitative$_{\!}$ results}$_{\!}$ on$_{\!}$ R4R~\cite{jain2019stay} (\S\ref{exp_comparison}).}}
    \label{table_R4R}
\vspace*{-15pt}
\end{table}

\begin{figure*}[t]		
	\begin{center}
		\includegraphics[width=0.98\linewidth]{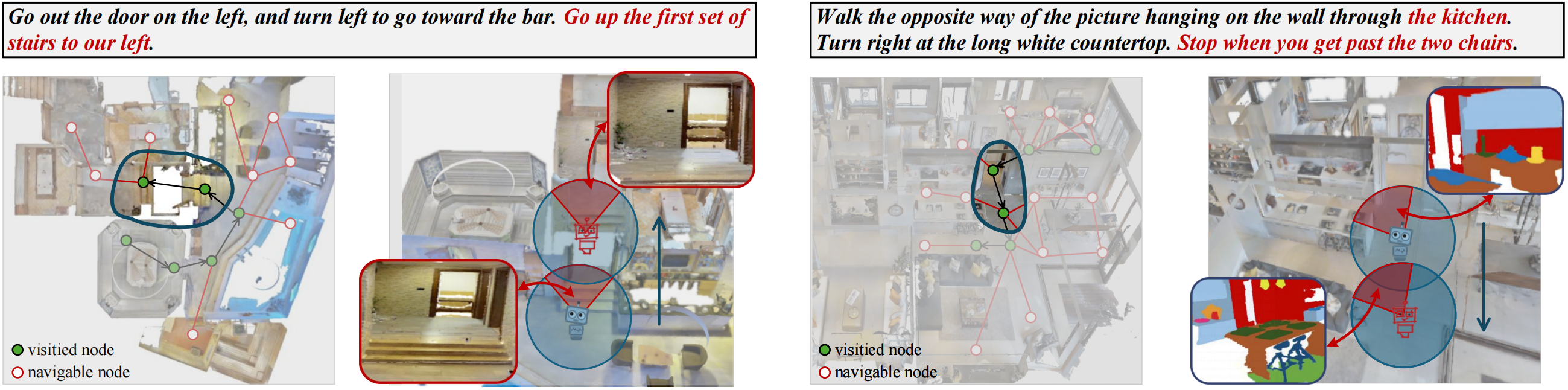} 
                \put(-371,-8){{(a)}} 
        \put(-122,-8){{(b)}}
        	\vspace{-6pt}
		\caption{\small{\textbf{Visualization of 3D Gaussian Maps} on R2R~\cite{AndersonWTB0S0G18} \textit{val unseen} split. Benefiting from the geometric priors and open-set semantics of the 3D Gaussian Map, our agent achieves a comprehensive understanding of spatial structures and semantic contexts. This enables our agent to (a) accurately interpret geometric transformations, such as \textit{``go up the first set of stairs"}, and (b) reason about fine-grained object relationships, as demonstrated by identifying and navigating around \textit{``the two chairs"}.} See \S\ref{exp_comparison} for more details.}
            \label{fig:add}
	\end{center}
	\vspace{-15pt}
\end{figure*}

\noindent\textbf{Performance on R4R~\cite{jain2019stay}.} R4R places higher demands on the capabilities of the agent in multi-stage reasoning and long-horizon planning. As shown in Table~\ref{table_R4R}, our method maintains a strong performance on R4R, consistently outperforming existing approaches. Specifically, compared to HAMT~\cite{chen2021history}, our approach achieves improvements of \textbf{2}\% in SR, CLS, and nDTW, with \textbf{3}\% gain in SDTW. These results further demonstrate the robustness of our method in maintaining spatial and semantic consistency on extended paths.

\noindent\textbf{Visual Results.} We conduct qualitative 
analysis to showcase the effectiveness of our approach. Fig.~\ref{exp_vis_ab} (a) depicts a case where the instruction requires the agent to navigate through multiple rooms and landmarks, such as the \textit{``bookcase"} and the \textit{``kitchen storage area"}, to reach the target location. This scenario requires the agent to accurately interpret both semantic cues and spatial relationships. The results show that our agent successfully identifies the intended path, whereas BEVBert~\cite{an2023bevbert} deviates to an incorrect room upon exiting the \textit{``bedroom"}. This demonstrates that our 3D Gaussian Map enables the agent to recognize and integrate semantic and geometric information from the environment, leading to more precise navigation. Moreover, Fig.~\ref{exp_vis_ab} (b) illustrates a scenario where the agent is required to navigate through constrained spaces and localize specific objects within a designated room, such as the \textit{``bathroom"} and the \textit{``rug"}. This task emphasizes fine-grained spatial reasoning and object-aware localization. Our agent precisely locates the target objects, while BEVBert~\cite{an2023bevbert} struggles to distinguish intricate spatial relationships in such narrow environments. This success highlights the advantage of our 3D Gaussian Map in capturing detailed scene information, thereby enabling the agent to achieve accurate navigation.

\begin{table}[t]
    \begin{center}
        \resizebox{0.49\textwidth}{!}{
            \setlength\tabcolsep{6.5pt}
            \renewcommand\arraystretch{1.05}
            \begin{tabular}{c|ccc|cc|ccc}
                \hline \thickhline
                \rowcolor{mygray}
                & \multicolumn{3}{c|}{Components} & \multicolumn{2}{c|}{R2R~\cite{AndersonWTB0S0G18}} & \multicolumn{3}{c}{REVERIE~\cite{qi2020reverie}} \\
                \cline{2-9}
                \rowcolor{mygray}
                \multicolumn{1}{c|}{\multirow{-2}{*}{\#}} & {ESM} & {OSG} & {MAP} & {SR}$\uparrow$ & {SPL}$\uparrow$ & {SR}$\uparrow$ & {RGS}$\uparrow$ & {RGSPL}$\uparrow$ \\
                \hline
                \hline
                1 & -- & -- & -- & 72 & 60 & 46.98 & 32.15 & 23.03 \\
                2 & \checkmark & -- & -- & 73 & 61 & 47.10 & 32.80 & 23.18 \\
                3 & \checkmark & \checkmark & -- & 75 & 64 & 50.50 & 34.83 & 24.75 \\
                4 & \checkmark & -- & \checkmark & 73 & 63 & 49.30 & 35.20 & 23.45 \\
                5 & \checkmark & \checkmark & \checkmark & \textbf{77} & \textbf{66} & \textbf{53.59} & \textbf{36.73} & \textbf{26.74} \\
                \hline
            \end{tabular}
        }
    \end{center}
    \vspace*{-12pt}
    \captionsetup{font=small}
    \caption{\small{\textbf{Ablation studies of the overall design} on \textit{val unseen} split of R2R~\cite{AndersonWTB0S0G18} and REVERIE~\cite{qi2020reverie}. See \S\ref{exp_abalation} for more details.}}
    \label{table:component_study}
    \vspace*{-8pt}
\end{table}

In addition, we highlight the strengths of our approach in both spatial and semantic understanding. In Fig.~\ref{fig:add} (a), we explicitly synthesize view-level 3D Gaussian Maps at different waypoints, showing that our method naturally encodes rich 3D spatial information, which previous methods lack. Based on these maps, the agent accurately interprets the geometric context to \textit{``go up the first set of stairs"}, illustrating how our method utilizes geometric priors to improve spatial awareness. Moreover, in Fig.~\ref{fig:add} (b), the agent navigates through a complex environment with multiple objects and intricate spatial relationships. Leveraging the 3D Gaussian Map, our agent successfully identifies key regions and objects, such as \textit{``the kitchen"} and \textit{``the two chairs"}, and perceives precisely about their spatial configuration. This indicates how our approach enables fine-grained semantic understanding, which in turn enhances VLN performance.

\begin{figure}[t]
	\begin{center}
		\includegraphics[width=0.98\linewidth]{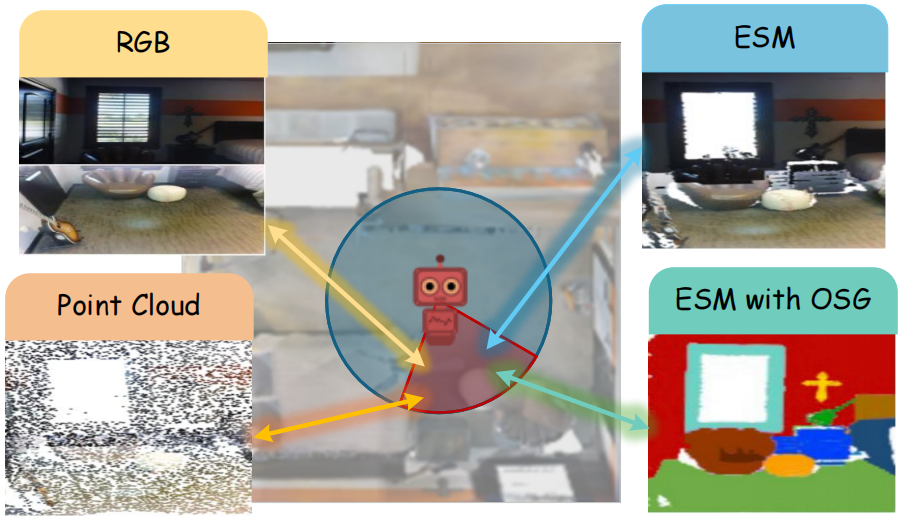}

	\end{center}
	\vspace{-7pt}
	\captionsetup{font=small}
   	\vspace{-10pt}
	\caption{\small{\textbf{Visualization of various scene map types on the same view.} Our method supports explicit visualization of 3D scenes, whereas previous methods are constrained to 2D rendered results. The visualization includes RGB images, 3D Point Clouds, Egocentric Scene Map (ESM, \S\ref{sec_EGM}), and ESM with Open-Set Semantic Grouping (OSG, \S\ref{sec_SSA}).} See \S\ref{exp_abalation} for more details.}
	\label{exp_abla}
	\vspace{-15pt}
\end{figure}

\subsection{Diagnostic Experiment}\label{exp_abalation} 
To evaluate each component, we conduct diagnostic studies on \textit{val unseen} splits of both R2R~\cite{AndersonWTB0S0G18} and REVERIE~\cite{qi2020reverie}.

\noindent\textbf{Overall Design (Fig.~\ref{fig_overview}).} We first assess the contributions of each component by progressively incorporating ESM (\S\ref{sec_EGM}), OSG (\S\ref{sec_SSA}), and MAP (\S\ref{sec_MAP}) into the baseline model (row \#1). As detailed in Table~\ref{table:component_study}, each module contributes incrementally to the performance. In particular, rows \#4 and \#5 highlight the impact of OSG (\textit{e.g.}, $73\%\!\rightarrow\!\textbf{77}\%$ for SR on R2R and $35.20\%\!\rightarrow\!\textbf{36.73}\%$ for RGS on REVERIE). Similarly, the comparison between rows \#3 and \#5 highlights the effectiveness of MAP (\textit{e.g.}, $75\%\!\rightarrow\!\textbf{77}\%$ for SR on R2R and $34.83\%\!\rightarrow\!\textbf{36.73}\%$ for RGS on REVERIE). Rows \#1 and \#5 show that combining all components together results in the largest gain over the baseline (\textit{e.g.}, $72\%\!\rightarrow\!\textbf{77}\%$ for SR on R2R and $32.15\%\!\rightarrow\!\textbf{36.73}\%$ for RGS on REVERIE).

\noindent\textbf{Analysis of ESM (\S\ref{sec_EGM}).} We next visually compare ESM with a conventional 3D point cloud (see Fig.~\ref{exp_abla}) to demonstrate its advantages. Unlike the sparse and noisy 3D point cloud, ESM constructs a spatially coherent map with fine-grained geometry. This improved map enhances the geometric awareness of the agent, assisting it in identifying spatial structure and accessible paths.

\noindent\textbf{Analysis of OSG (\S\ref{sec_SSA}).} We further investigate the impact of OSG. Fig.~\ref{exp_abla} visualizes the 3D scene synthesized by ESM with OSG. The results show that Gaussians in ESM are grouped according to their object instance or stuff membership in the 3D scenario, demonstrating that OSG injects enriched semantics into ESM while ensuring cross-view consistency. This open-set semantics enhances the agent’s ability to infer object relationships and scene structures, thereby improving its decision-making in VLN.

\noindent\textbf{Analysis of MAP (\S\ref{sec_MAP}).} To assess the contributions of Scene, View, and Instance levels, we evaluate models with different level combinations. From Table~\ref{table:MAP_levels}, we can observe that: \textbf{\textit{i)} Row \#1 $_{\!}$\textit{vs} $_{\!}$\#2 $_{\!}$\textit{vs} $_{\!}$\#3 $_{\!}$\textit{vs} $_{\!}$\#4:} Each level contributes to performance gain, and the Instance level providing the most significant boost (\textit{e.g.}, $72\%\!\rightarrow\!\textbf{74}\%$ for SR on R2R). \textbf{\textit{ii)} Row \#1 $_{\!}$\textit{vs} $_{\!}$\#2 $_{\!}$\textit{vs} $_{\!}$\#5 $_{\!}$\textit{vs} $_{\!}$\#6:} Combining multiple levels yields further enhancements and the best results are achieved when all levels are integrated  (\textit{e.g.}, $72\%\!\rightarrow\!\textbf{77}\%$ for SR on R2R), which indicates their complementarity.

\begin{table}[t]
    \begin{center}
        \resizebox{0.49\textwidth}{!}{
            \setlength\tabcolsep{6.5pt}
            \renewcommand\arraystretch{1.05}
            \begin{tabular}{c|ccc|cc|ccc}
                \hline \thickhline
                \rowcolor{mygray}
                & \multicolumn{3}{c|}{MAP Levels} & \multicolumn{2}{c|}{R2R~\cite{AndersonWTB0S0G18}} & \multicolumn{3}{c}{REVERIE~\cite{qi2020reverie}} \\
                \cline{2-9}
                \rowcolor{mygray}
                \multicolumn{1}{c|}{\multirow{-2}{*}{\#}} & {Scene} & {View} & {Instance} & {SR}$\uparrow$ & {SPL}$\uparrow$ & {SR}$\uparrow$ & {RGS}$\uparrow$ & {RGSPL}$\uparrow$ \\
                \hline
                \hline
                1 & -- & -- & -- & 72 & 60 & 46.98 & 32.15 & 23.03 \\
                2 & \checkmark & -- & -- & 73 & 63 & 48.53 & 33.61 & 23.50 \\
                3 & -- & \checkmark & -- & 73 & 61 & 47.21 & 33.78 & 22.76 \\      
                4 & -- & -- & \checkmark & 74 & 62 & 49.32 & 35.42 & 24.12 \\
                5 & \checkmark & \checkmark & -- & 74 & 64 & 51.64 & 34.17 & 24.00 \\
                6 & \checkmark & -- & \checkmark & 75 & 64 & 52.42 & 35.64 & 24.57 \\
                7 & \checkmark & \checkmark & \checkmark & \textbf{77} & \textbf{66} & \textbf{53.59} & \textbf{36.73} & \textbf{26.74} \\
                \hline
            \end{tabular}
        }
    \end{center}
    \vspace*{-13pt}
    \captionsetup{font=small}
    \caption{\small{\textbf{Ablation studies of MAP strategy} on \textit{val unseen} split of R2R~\cite{AndersonWTB0S0G18} and REVERIE~\cite{qi2020reverie}. See \S\ref{exp_abalation} for more details.}}
    \label{table:MAP_levels}
    \vspace*{-8pt}
\end{table}

\vspace*{8pt}

\section{Conclusion}
\label{sec:conclusion}
In this work, we propose a unified 3D Gaussian Map that integrates geometric priors with open-set semantics to enhance sequential decision-making in Vision-and-Language Navigation. Our agent first introduces the Egocentric Scene Map to project 2D panoramic observations into structured 3D representations that preserve geometric context. It then leverages the Open-Set Semantic Grouping operation to group these 3D primitives according to their context-aware semantic information. Finally, it adopts the Multi-Level Action Prediction strategy to refine navigation decisions by aggregating cues across scene-level layouts, view-specific features, and fine-grained instance-level semantics. Extensive qualitative and quantitative experiments demonstrate consistent improvements in navigation performance and validate the effectiveness of our method.

{
    \small
    \bibliographystyle{ieeenat_fullname}
    \bibliography{main}
}

\end{document}